# Intelligent architectures for robotics: The merging of cognition and emotion

Luiz Pessoa
Department of Psychology
Department of Electrical and Computer Engineering
Maryland Neuroimaging Center
University of Maryland, College Park, MD 20742, USA
pessoa@umd.edu

Abstract
What is the place of emotion in intelligent robots? In the past two decades, researchers have advocated for the inclusion of some emotion-related components in the general information processing architecture of autonomous agents, say, for better communication with humans, or to instill a sense of urgency to action. The framework advanced here goes beyond these approaches and proposes that emotion and motivation need to be integrated with all aspects of the architecture. Thus, cognitive-emotional integration is a key design principle. Emotion is not an "add on" that endows a robot with "feelings" (for instance, reporting or expressing its internal state). It allows the *significance* of percepts, plans, and actions to be an integral part of all its computations. It is hypothesized that a sophisticated artificial intelligence cannot be built from separate cognitive and emotional modules. A hypothetical test inspired by the Turing test, called the Dolores test, is proposed to test this assertion.

Highlights
- Most information-processing architectures of intelligent robots focus on cognition
- Cognitive-emotional integration is as a key design principle of autonomous agents
- Emotion is not an "add on" that endows a robot with "feelings"
- Emotion allows *significance* of percepts/actions to be part of robots' computations
- *Dolores test* evaluates if autonomous agents can be built without emotion

## 1. INTRODUCTION

In the first episode of Westworld, a popular HBO series, a human is examining Dolores, one of the central characters of the show. Dolores speaks with a heavy frontier accent. The human says "lose the accent," to which she promptly acquiesces. Her answers are charged with emotion, they convey a deep feeling of confusion and anxiety; she's distraught. The examiner then commands: "cognition only." Once again, Dolores does as she's told, losing all trace of emotion. Dolores is a "host" – how the humanoid robots are called – at Westworld, an Old West theme park created for the entertainment of human "guests" who are free to kill or have sex with them.

    For over two millennia, Western thinkers have separated emotion from cognition, emotion being the lowly sibling of the two. Cognition helps explain the quantum nature of matter and sends humans to the moon. Emotion might save the lioness in the savannah but makes humans act irrationally with

disconcerting frequency. Against this backdrop, how should autonomous robots be built? Should emotion be part of their information-processing architecture? If so, how should it interact with cognition?

Cognition refers to processes such as memory, attention, language, problem solving, and planning. Many cognitive processes are thought to involve sophisticated functions that might be uniquely human. Furthermore, they often involve so-called controlled processes, such as when the pursuit of a goal needs to be protected from distraction. Whereas there is relative agreement about what constitutes cognition, there is less accord about emotion. Some investigators link it to the concepts of drive and motivation, suggesting that emotions are states elicited by rewards and punishers [1]. Others propose that emotions are involved in the evaluation of events, called "appraisals" [2], which can be explicit or implicit. Some approaches focus on basic emotions (for example, fear and anger [3]) while others focus on an extended set of emotions, including moral ones (for example, pride and envy [4]). Finally, many researchers suggest that emotions are strongly linked to bodily states [5].

In building an "intelligent" robot it might seem best to focus on its cognitive capacities, perhaps disregarding emotion entirely, or perhaps including only as much of emotion as necessary. Fig. 1A illustrates this scenario, where the blue segment represents a cognitive module and the orange segment indicates the emotion module of a hypothetical robot. When emotion is included, to avoid "emotional interference," another design property would allow the cognitive module to downregulate emotion inputs, minimizing their influence to the desired extent. Thus, emotion is restricted to contributing to behavior only when deemed useful by the cognitive module.

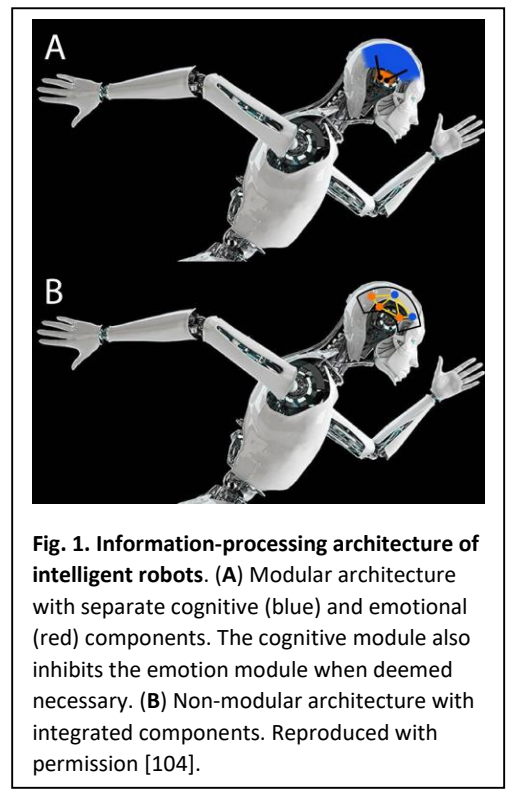

It is conceivable that the robot in Fig. 1A could function effectively in some settings. The cognitive architecture would endow it with the required problem-solving capabilities, and the emotion module would instantiate a sense of urgency into decisions and actions [6, 7]. The argument developed here, however, is that this type of architecture, in general, is inherently limited in its behavioral capabilities. For sophisticated intelligent behavior, I argue that cognition and emotion need to be intertwined in the general information-processing architecture (Fig. 1B).

What is meant by "sophisticated intelligent behavior?" While the goal here is not to precisely answer this question, the question can be rephrased as follows: "Does building models of emotional/affective mechanisms into our robots make them more natural, more useful, or more efficient?" [8]. Furthermore, a chief target goal should be *autonomy*, that is, the ability to function independently in diverse and challenging conditions and environments [9] (for example, interplanetary travel). Importantly, the argument made here for a cognitive-emotional architecture is not simply rooted in the need for better communication with humans ([10]; given that the latter do have emotion) or, relatedly, to generate emotional expressions [10], or even for survival-related purposes [7]. The contention is that, for the types of intelligent behaviors frequently described as cognitive (say, attention, problem solving, planning), the integration of emotion and cognition is necessary [11].

**Fig. 1. Information-processing architecture of intelligent robots**. (**A**) Modular architecture with separate cognitive (blue) and emotional (red) components. The cognitive module also inhibits the emotion module when deemed necessary. (**B**) Non-modular architecture with integrated components. Reproduced with permission [104].

To motivate the information-processing architecture described here, I will first describe growing evidence that emotion and cognition are integrated in the brain. Next, general concepts of brain evolution will be reviewed, and implications for the organization of emotion and cognition in the brain discussed. Based on these ideas, a cognitive-emotional architecture will be motivated and proposed properties described. I will then develop the hypothesis that a humanoid robot cannot exhibit fully autonomous "interesting" behaviors without cognitive-emotional integration, which will be summarized via the "Dolores test." General conclusions will be then outlined.

**2. COGNITION AND EMOTION IN THE BRAIN**

How is emotion organized in the brain? Historically, textbooks have described the emotional brain by listing a series of brain regions and discussing their purported main function(s) (see also [12]). Several subcortical regions would be featured, including the amygdala ("fear processing"), hypothalamus ("survival mechanisms"), periaqueductal gray ("defensive behaviors"), hippocampus ("emotional memory"), nucleus accumbens ("reward processing"), and so on. Cortical regions would include the cingulate cortex ("affective action"), insula ("processing of bodily signals"), and orbitofrontal cortex ("reward processing").

Although the idea that some brain regions are specialized for emotion is favored by some researchers [13], a different view has emerged in the past two decades, one in which emotion and cognition strongly interact and, in fact, are integrated in the brain [11, 14]. Before describing this perspective, I will first motivate it in intuitive terms.

Consider a hypothetical organism that has its visual cortex (the part that receives signals from the eye's retina) and its auditory cortex (the part that receives signals from the ear's cochlea) heavily interconnected with each other. This architecture means that every time signals are processed by the visual cortex, they strongly influence auditory cortex, and vice versa – vision and audition mutually influence each other. Indeed, it would be better to call the parts of the brain in question visual-auditory (or audio-visual) areas. Even without studying behavior in our hypothetical organism, we could make the strong prediction that it would be unable to respond in a purely visual or auditory way. This is because its brain is built such that the two sensory domains have, in effect, no clear boundary – they are not independent modules.

Emotion and cognition are not separable in the human brain for similar reasons. Knowledge about anatomy and physiology indicates that the two are closely intertwined [15]. Next, I describe principles of brain organization that reveal why emotion is not a distinct unit of the brain.

## 2.1. Massive Combinatorial Anatomical Connectivity

Computational analysis of anatomical connectivity demonstrates that both cortical and subcortical brain regions are densely interconnected [16-20] (Fig. 2). Rich connectivity is not limited to specific sectors of the brain (say, prefrontal cortex) but instead encompasses all of them (including brainstem and cerebellum). Central anatomical properties include: 1) Massive interconnectivity; 2) High global *accessibility* [21]; and 3) The existence of a "connectivity core" or "rich-club" of regions marked by especially high levels of connectivity [21, 22]. Focusing on macaque cortical regions, one study described a core set of 17 regions spanning parietal, temporal, and frontal cortex that was marked by 92% connectivity density (92% of the connections that could exist were present) [23]. By combining multiple sources of data, another study described a core that was distributed across all major brain sectors (all cortical lobes, thalamus, and subcortical regions in the forebrain) [22].

In all, the anatomical connectivity is such that signals can flow between disparate regions in very few steps. And because *hub* regions (those that are particularly well connected) aggregate and distribute information widely, diverse types of signals (say, related to perception and action) have the opportunity to influence one another [24].

## 2.2. High Distributed Functional Connectivity

Understanding brain function requires characterizing how regions are "functionally connected." The idea of *functional connectivity* was devised to characterize how neurons (or regions) interact and was defined as the "temporal coherence" among their activity, for example, as measured by correlating their signals [25-27]. Hence, functional connectivity is essentially a description of the *joint state* of two brain elements, regardless of their anatomical connectivity status (linked by a strong or weak anatomical connection, or even not directly linked at all).

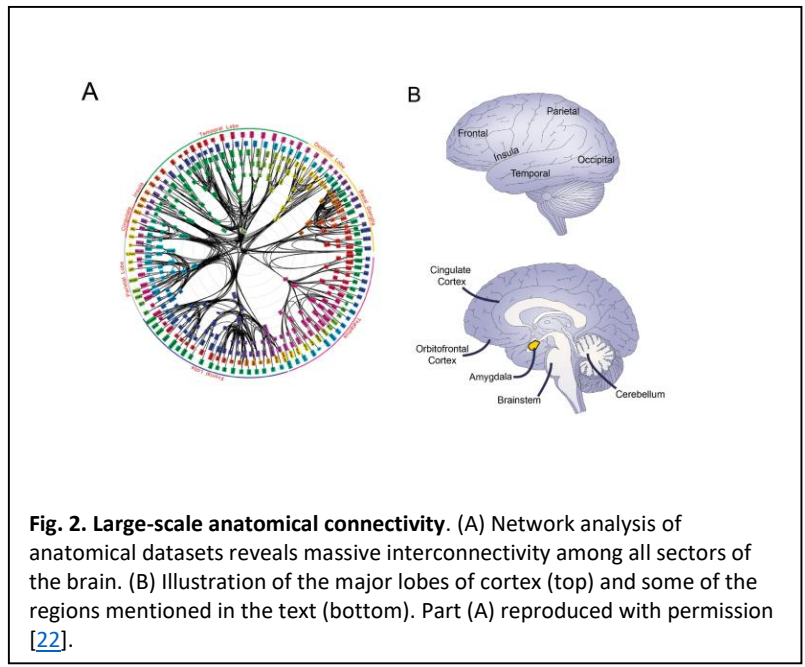

**Fig. 2. Large-scale anatomical connectivity**. (A) Network analysis of anatomical datasets reveals massive interconnectivity among all sectors of the brain. (B) Illustration of the major lobes of cortex (top) and some of the regions mentioned in the text (bottom). Part (A) reproduced with permission [22].

Thus, in general, what is important is not simply a region's anatomical location, but its position in a space of functional relationships to other regions [27-29]. From the perspective of a brain region, at a given time, a region affiliates with a set of other regions, thereby defining a momentary functional circuit. And relatively spared coordinated activity can emerge in brains with substantially altered structural connectivity [30, 31]. In all, anatomical architectural features support the efficient communication of information even when strong direct structural connections are absent [32-34], and allow functional interactions that vary as a function of context.

It should be pointed out that "functional connectivity" does not distinguish certain scenarios. Consider, for example, two neuronal populations, A and B[1]. Even if A and B are temporally coherent in their activation but only one triggers a causally related consequence in C, the functional connectivity between A and B does not include that information (and some could even question the validity of stating that they are functionally connected). Whereas this is potentially a problem, establishing causation in complex systems is far from straightforward. In fact, the type of Newtonian "billiard ball" causation model at times applied to reasoning about the brain (in this case, A "causes" B much like a billiard ball hitting another) may be in need of a revision [35-37]. For example, "probabilistic causation" [38] or "dynamical systems causation" offer promising avenues. In the latter, the goal is to investigate causation in (possibly non-linear) dynamical systems, and two variables are "causally linked" if they participate in the same dynamical system [39].

## 2.3. Overlapping and Dynamic Brain Networks

Networks of brain regions collectively support behaviors. Thus, the *network itself is the unit*, not the brain region. Indeed, research characterizing both anatomical (e.g., [22]) and functional (e.g., [40]) networks has grown rapidly in the last decade, and a host of specific networks have been proposed, such as the "salience" (important during the processing of salient events) and "executive" (important during cognitive processes, including attention and working memory) networks [41].

---
[1] I thank a reviewer for bringing up this point.

At the same time, analysis of anatomical/functional data has repeatedly revealed regions with widespread connectivity (see section above). This implies that describing brain network in terms of disjoint sets of regions (where each brain region belongs to a single network or cluster) only partly captures the complexity of the organization. It is important, therefore, to elucidate the *overlapping* structure of networks [42]. In several scientific disciplines, there is increasing realization that actual networks are best understood in terms of overlapping subsystems [43]. In neuroscience, disjoint networks (that is, non-overlapping) have been favored, although recent investigations have started to study potential ways to characterize overlapping organization [44, 45].

In a recent study, we investigated the potential overlapping structure of functional brain connectivity in a data-driven fashion [46]. Instead of assuming that regions belong to a single network, we allowed them to belong to *all* communities (Fig. 3). Thus, when community organization was determined, a node's *participation* to every community was estimated; participation strengths were called *membership values*. Each node was assigned a probability-like membership value to each of the existing communities, resulting in a membership vector with entries between 0 and 1 (and summing to 1); entries close to 1 indicated membership to essentially one community, that is, the non-overlapping case. The approach to detecting overlapping communities we adopted was originally based on the mixed-membership stochastic blocks model [47] within the context of stochastic variational inference [48]. The specific algorithm we employed was developed by Gopalan and Blei [49], and can be viewed as a generative Bayesian algorithm favoring assortativity (nodes with similar memberships are more likely to cluster together).

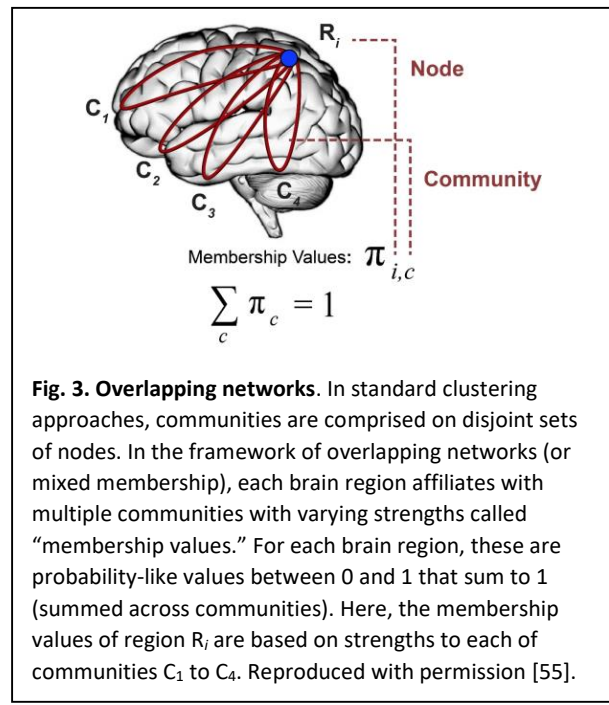

Fig. 3. Overlapping networks. In standard clustering approaches, communities are comprised on disjoint sets of nodes. In the framework of overlapping networks (or mixed membership), each brain region affiliates with multiple communities with varying strengths called "membership values." For each brain region, these are probability-like values between 0 and 1 that sum to 1 (summed across communities). Here, the membership values of region $R_i$ are based on strengths to each of communities $C_1$ to $C_4$. Reproduced with permission [55].

When applied to resting-state fMRI data, the algorithm captured community organization matching many of the general features of standard disjoint networks. However, the analysis indicated that overlapping communities are not fully captured by a single disjoint community, and that they contain information that is not well described by disjoint communities. To gain a richer understanding of the overlapping networks obtained, for each community, we considered the entire distribution of membership values, that is, we considered the membership values of all brain regions simultaneously. A community containing nodes that affiliated mostly with that community would exhibit a membership distribution with values close to 1. Instead, we found that all mixed communities had many regions with non-trivial membership values far from the peak of 1, which indicated that the overlapping communities were not well described by "exclusive nodes" (nodes that participated in only one community) but by "multi-partner nodes" (nodes that participated in multiple communities).

Furthermore, brain networks are not static but evolve temporally [50, 51]. Although anatomical pathways change across the life span, faster dynamics are an essential ingredient of functional connections between regions, which vary as a function of context, and are altered by cognitive, emotional, and motivational variables, among others. Therefore, network organization is a *dynamic* property. Indeed, the growth of methods to characterize time-varying functional connectivity has begun to yield novel characterizations of how network organization evolves [52-54]. In sum, networks are not static and fixed collections of brain regions; they are dynamic coalitions of regions that form and dissolve to meet specific computational needs. This poses several challenges, as the very notion of a network as a coherent unit is undermined. For instance, at what point does a coalition of regions become something *other* than, say, the salience network? An approach to modeling networks in general that partly mitigates this problem is to conceptualize networks as inherently overlapping. Instead of treating nodes as belonging to a single network, in overlapping networks, each node is a member of every network with a specific probability like membership value, which can then fluctuate across time [55].

Dynamic aspects of network function naturally will benefit from understanding interactions in terms of non-linear dynamical systems. Such systems undergo qualitative changes in behavior (so-called bifurcations) as a function of their input and/or parameter changes. For example, Hansen and colleagues [56], show via simulations of a whole-brain mean-field computational model that functional connectivity exhibits rapid transitions between a few discrete states. In the so-called "resting-state" (when participants are not explicitly engaged in a task), such state transitions are spontaneous and often very sharp.

## 2.4. Anatomical Connectivity to and from the Body

Emotion is closely linked to bodily states, and the very history of emotion research is defined by the discussion and controversy surrounding the question of the necessity of the body for emotion [5]. The exact role of the body in emotion notwithstanding, brain circuits that involve signals to and from the body are important for associated processes.

Brain connectivity affecting the body originates in both cortex and subcortex. In cortex, a notable region is the cingulate cortex [57,58], whose descending projections to autonomic regulatory structures have been amply described (autonomic functions include respiration, heart rate, and pupil response) [59]. This connectivity is consistent with effects of cingulate electrical stimulation on virtually all autonomic processes, as well as many endocrine mechanisms. Conversely, visceral pathways carrying signals from the body to the brain reach multiple subcortical and cortical regions. In particular, cortical regions with notable body-related signals include the orbitofrontal cortex, the cingulate gyrus, and the insula [60]. For example, the physiological condition of the entire body is conveyed to the posterior insular cortex [60], which can be considered interceptive cortex, much like parts of parietal cortex are somatosensory cortex, for example. In conclusion, both cortex and subcortex are part of extensive connectional systems that link the body to the brain. In this manner, the entire spectrum of brain signals can affect the body, and vice versa.

## 2.5. Merging Cognition and Emotion in the Brain

In summary, anatomical and functional studies reveal that brain regions are massively interconnected. In particular, the brain basis of emotion involves large-scale cortical-subcortical networks that are distributed and sensitive to bodily signals [15]. The high degree of signal distribution and integration provides a nexus for the intermixing of information related to perception, cognition, emotion, motivation, and action. Importantly, the functional architecture consists of multiple overlapping networks that are highly dynamic and context sensitive. Thus, how a given brain region affiliates with a specific network shifts as a function of task demands and brain state.

Whereas some regions are more highly interconnected than others, overall information can travel from one part of the brain to another in just a few steps. This implies that the brain's architecture does not leave much room for isolation of signals; it is promiscuous, with information of a given type being mixed with others. Furthermore, adopting a large-scale network perspective to brain organization helps clarify why some brain structures, such as the amygdala, are thought to be important for emotion: they are important hubs of large-scale connectivity systems. It also illuminates why the impact of emotion is so wide ranging – for one, it is not possible to impact emotion without affecting cognition.

### 3. BRAIN EVOLUTION, COGNITION, AND EMOTION

From an evolutionary perspective, it is sometimes stated that cognition can be considered "the tip of the iceberg," often associated with "newer" parts of the cortex, while the rest of the iceberg involves emotional processes that rely on "old" subcortical brain regions. Furthermore, it is presumed that the old emotional system functions in a largely autonomous manner (for discussion, see [61]).

Some brain system can indeed be considered old. For example, anatomical and behavioral data suggest that ray-finned fishes and land vertebrates probably share an ancestor (>420 million years ago) that already possessed some form of an amygdala [62] – a subcortical region that is believed to be particularly important for emotional processing [63]. And the amygdala of the ancestral amniote (which includes reptiles, birds, and mammals; around 340 million years ago) is believed to have had parts that are key subregions of the human amygdala, for example [62].

In the beginning of the twentieth century, neuroanatomists proposed the idea that brain evolution takes place in a fairly progressive fashion, from fishes to amphibians, to reptiles, to birds, and then mammals (culminating with the human brain) – the idea of progression goes back to Aristotle. In this scheme, the brain evolved via the *sequential addition* of parts. Thus, in the brain of a cat (a mammal) one could discern older parts that remained largely unchanged underneath the cortex [64]. Furthermore, it was proposed that structures at the top control the ones at the bottom [61]. Unfortunately, this erroneous viewpoint [65] became dominant in neuroscience for many decades. It reached its most dramatic example in Paul MacLean's proposal of the "triune brain," which consisted of a reptilian complex (at the base of the brain, including brainstem), to which an "old brain" was added, and later a "new cortex" included in mammals [66]. In the context of the present review, it is noteworthy that the organization shown in Fig. 1A very much resembles this notion.

It is now known that such framework does not reflect current knowledge of comparative neuroanatomy and does not adequately describe brain evolution (not even in general terms) [65,67]. But what is the *basic plan* of the vertebrate brain? The chassis of the brain of fishes, amphibians, and all other vertebrates contains both subcortex (with multiple parts) and simple forms of cortex (called pallium). With evolution, brain changes occurred across the brain. But importantly, new circuits originated in a manner that affected both cortex and subcortex.

An example of this type of *mutual embedding* is provided by the amygdala. In mammals, this structure is composed of a dozen or more subregions. In one study it was found that some amygdala subregions are considerably more "developed" in monkeys than in rats (based on morphological characteristics, such as cell counts and the volume of subregions) [68]. One possibility is that the observed differences between rats and monkeys are linked to the degree of connectivity of amygdala subregions to other brain structures (for related discussion, see [67]). In other words, the changes in the amygdala may have paralleled the development of the cortical areas with which these nuclei are interconnected in primates.

Taken together, brain evolution likely proceeded via the mutual embedding of new and old structures, not by the modular addition of parts. Furthermore, separation of "old" and "new" parts does not take into account the extensive anatomical connectivity of brain circuits, as previously discussed, nor the overlapping and dynamic nature of brain networks.

## 4. COGNITIVE-EMOTIONAL ARCHITECTURES

With the discussion of the previous two sections in mind, we can return to the central question addressed here: What kind of architecture should an intelligent, autonomous robot have? Although a broad range of cognitive architectures have been proposed [69-73]; for review, see [74]), a general principle is that they are modular (Fig. 4A). For example, they may have low-level modules for action and perception, mid-level modules for task coordination and mid-level perception, and higher levels for task planning and visual cognition, say. Frequently, the top level is connected to a knowledge database. To coordinate the modules, a central controller or supervisor is also posited, which utilizes temporary buffers as part of an active workspace. By and large, the architecture is hierarchical, although parallel components and organization are recognized as important, too [75].

In the past two decades, a steady stream of researchers have advocated for the inclusion of emotion-related components to the general information processing architecture (or "cognitive" architecture, as sometimes it is still referred) [10, 70, 76-82]; see contributions in [83]). One type of argument is that emotion components are needed to instill urgency to action and decisions [7]. Others have advocated emotion components to aid understand emotion in humans, or to generate human-like expressions (for empathy, say) [10]. In broad terms, a first-pass effort at including affect entails adding an emotion module so that it can influence other components of the architecture (Fig. 4B). Another aspect emphasized in the literature is that emotion can be used to generate an internal state that influences actions [e.g., 82, 84] (Fig. 3B, right).

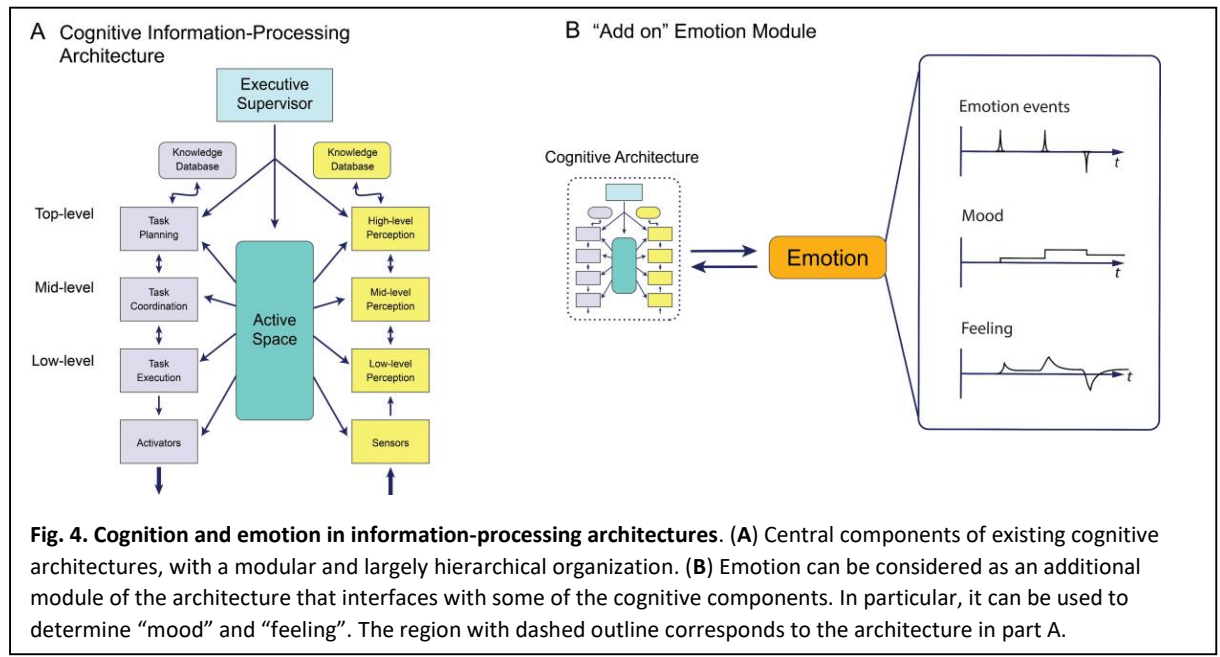

**Fig. 4. Cognition and emotion in information-processing architectures**. (**A**) Central components of existing cognitive architectures, with a modular and largely hierarchical organization. (**B**) Emotion can be considered as an additional module of the architecture that interfaces with some of the cognitive components. In particular, it can be used to determine "mood" and "feeling". The region with dashed outline corresponds to the architecture in part A.

The framework advanced here goes beyond these approaches and proposes that emotion and motivation need to be integrated with all aspects of the architecture. In particular, emotion-related mechanisms influence processing beyond the modulatory aspects of "moods" linked to internal states (hunger, sex-drive, etc.). Of note, in the literature, emotion is often linked to evaluative aspects of the organism-environment relationship and with affective states; motivation is associated with how the organism acts in a given situation (approach vs. withdrawal). Given the integrationist framework described here, I do not make this distinction [11].

As described previously, emotion is not neatly encapsulated in a specific part of the brain – one cannot point to emotion in the brain. In addition, cognition was not a late addition to the mammalian, or human, brain [65]. So, how can we operationalize it in an information-processing architecture? Emotion can be thought of as a set of *valuating mechanisms* that help organize behavior [85], for instance by helping take into account both the costs and benefits linked to percepts and actions [86]. At a general level, it can be viewed as a biasing mechanism, much like the "cognitive" function of attention. However, such conceptualization is overly simplistic because emotion does not amount to just providing an extra boost to a specific sensory input, potential plan, or action. When the brain is conceptualized as a complex system of highly interacting networks of regions, we see that emotion is interlocked with perception, cognition, motivation, and action. Whereas we can refer to certain behaviors as "emotional" or "cognitive," this is only a language short-cut. Thus, the idea of a biasing mechanism is too limited. From a design perspective, all components of the architecture should be influenced by emotional and motivational variables (and vice versa). Thus, the architecture must be strongly *non-modular*.

To illustrate these ideas, I will describe architectural features inspired by knowledge of brain and behavior. The focus will be on visual perception and executive control, but related schemes can be used in the case of action, too, as adaptive behavior requires monitoring and evaluating the outcome of actions. I will build upon a conceptual framework called the *dual competition model* which describes how both emotional and motivational signals are integrated with perception and cognition so as to effectively incorporate *value* into the unfolding of behavior [87]. The general framework is displayed in Fig. 5A. Because the use of boxes invites a modular interpretation, the architecture is better represented in terms of overlapping networks of processes or mechanisms (Fig. 5B).

## 4.1. Perceptual competition

Objects in the environment compete for limited perceptual processing capacity and control of behavior [88-90]. Because processing capacity for vision is limited [91], selective attention to one part of the visual field comes at the cost of neglecting other parts. Thus, a popular notion is that there is *competition* for resources in visual cortex [88,92].

In operationalizing competition, we can use the concept of a *priority map*, which contains representations of spatial locations that are behaviorally important [93,94]. Traditionally, "bottom-up" factors (such as stimulus salience) and "top-down" factors (such as goal relevance) were emphasized as the major inputs to determine priority. To these, we must add affective significance (for instance, based on the prior co-occurrence of a stimulus with aversive events) and motivational significance (for instance, based on the prior co-occurrence of a stimulus with rewards) (Fig. 6A) (see also [95]). In terms of implementation, the architecture requires mechanisms to embed affective and motivational significance, as well as stimulus- and goal-related factors, into perception. This can be accomplished via an explicit priority map (Fig. 6A) or by having multiple influences converge on representations of sensory stimuli (Fig. 6B). In the brain, perceptual competition occurs in part via interactions between regions important for attention (in frontal and parietal cortices, for example) and those that are particularly tuned to the evaluation of affective and motivational significance.

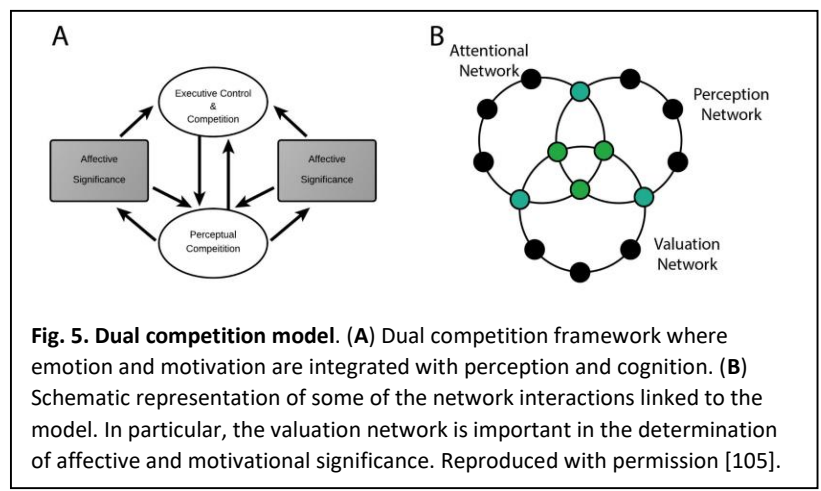

**Fig. 5. Dual competition model**. (**A**) Dual competition framework where emotion and motivation are integrated with perception and cognition. (**B**) Schematic representation of some of the network interactions linked to the model. In particular, the valuation network is important in the determination of affective and motivational significance. Reproduced with permission [105].

But the above should not be viewed as just an extension of the priority map with just additional variables. In particular, if one considers a predictive coding framework [96, 97], internal models of the current input will incorporate features along multiple dimensions, including sensory ones but also affective and motivational significance. Thus, establishing a predictive code involves building internal models that incorporate all of these dimensions simultaneously.

## 4.2. Executive control

At the heart of cognition, executive control refers to operations involved in maintaining and updating information, monitoring conflict and/or errors, resisting distracting information, inhibiting prepotent responses, and shifting mental sets. A useful way to conceptualize executive control is in terms of a set of processes, including *inhibition*, *updating*, and *shifting* [98]. These processes are not independent, however; when a given function is necessitated, resources devoted to one component will not be available to other operations and interference will ensue, possibly compromising performance [11].

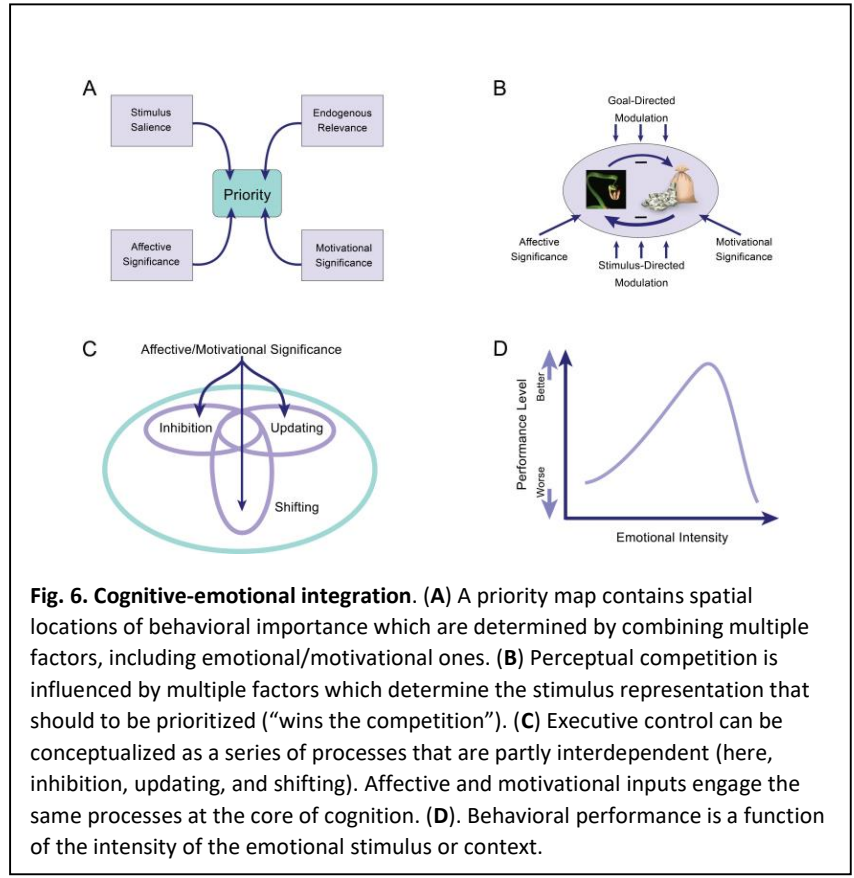

**Fig. 6. Cognitive-emotional integration**. (**A**) A priority map contains spatial locations of behavioral importance which are determined by combining multiple factors, including emotional/motivational ones. (**B**) Perceptual competition is influenced by multiple factors which determine the stimulus representation that should to be prioritized ("wins the competition"). (**C**) Executive control can be conceptualized as a series of processes that are partly interdependent (here, inhibition, updating, and shifting). Affective and motivational inputs engage the same processes at the core of cognition. (**D**). Behavioral performance is a function of the intensity of the emotional stimulus or context.

Dealing with an emotional stimulus or situation requires the types of behavioral adjustments that characterize executive function. For example, *updating* might be needed to refresh the contents of working memory, *shifting* might be recruited to switch the current task set, and *inhibition* might be invoked to cancel previously planned actions. In this manner, specific resources are coordinated in the service of emotional processing (Fig. 6C) and, if temporarily unavailable to additional task requirements, would compromise performance – the stronger the emotional manipulation, the stronger the interference (see below). A laboratory example might help illustrate these ideas. Suppose a participant is performing an effortful task of some sort and a change of background color signals that she will receive a mild shock sometime in the next 30 seconds. The participant may *update* the contents of working memory to include the "shock possible" information. *Shifting* may occur in that the participant may shift between the execution of the cognitive task and "monitoring for shock" (and possibly preparing for the aversive event) every few seconds. Now, if another cue stimulus indicated that shock would be delivered in the next second, the participant might temporarily *inhibit* response to the task to prepare for the shock. Altogether, dealing with the emotional situation necessitates the same types of executive functions considered to be the hallmark of cognition.

Importantly, the intensity of the emotional stimulus (or context) determines if it will improve or hinder behavioral performance. Moderate intensity, in particular, allows resources to be devoted to the situation at hand, improving performance. However, if the intensity is high enough, processes may be temporarily unavailable to handle the aspects of the situation adequately. This inverted-U type of relationship is common in many behavioral settings (Fig. 6D).

## 5. THE DOLORES TEST FOR ROBOTICS

Let us consider the humanoid robots of Westworld again. In the scene previously referred to, the central humanoid character, Dolores, was asked to consider and describe her situation by using cognition only. But could a human "lose all emotion" and describe complex events without a trace of affect like Dolores did? The present framework indicates that the answer to this question is "no." Unlike Dolores, humans cannot lose all emotion and simply proceed cognitively. The brain is organized such that the mind doesn't behave in this manner.

Dolores is not human; she is a robot built for the amusement of humans. Could a sophisticated artificial intelligence be built with separate cognitive and emotional modules? I contend that such a humanoid would be a far cry from the complex hosts of the series, whose behaviors are only possible when emotion and cognition are intertwined. Dolores, for one, would be unable to reflect about her past in a meaningful way just cognitively; emotion is part and parcel of her understanding of the world. And this is not just because she lives in a violent, ruthless world, but because making sense of reality necessitates the integration of emotion and cognition. Could this prediction be tested?

In 1950, Alan Turing [99] proposed a test to establish the presence of thought in a computer by assessing a machine's ability to exhibit intelligent behavior indistinguishable from that of a human. In the Turing test, an interrogator is supposed to be in a room separated from a machine and another person. The object of the game is for the interrogator to determine which of the other two is the person and which is the machine. The game is such that the human interrogator is physically separated from the other two. I propose, instead, the *Dolores test*: one cannot build a humanoid that behaves in a manner as sophisticated as Dolores does that is built with separate emotional and cognitive modules.

A scenario somewhat similar to that envisioned by the Dolores test is featured in the 1982 movie Blade Runner. Humans paid to track down rogue humanoids use the "Voight-Kampff device," a polygraph-like machine used to help determine whether or not an individual is a robot. The machine measures bodily signals linked to respiration, blush response, heart rate, and eye movement in response to questions dealing with emotion and empathy. However, the Dolores test proposed here is fundamentally distinct from the one in Blade Runner. It is not based on the idea that bodily responses to emotionally charged situations would deviate in a detectable way in robots. Instead, it is based on the hypothetical scenario that a robot could be built and that it would display "sophisticated behaviors" of the kind produced by the robots in Westworld. The idea of the test is that, by observing behaviors as the humanoids interact with both other humanoids and humans, humans and humanoids would *not* be confused with each other if the latter were built according to a modular cognitive-emotional architecture (Fig. 7) – that is, complex behaviors require a non-modular architecture.

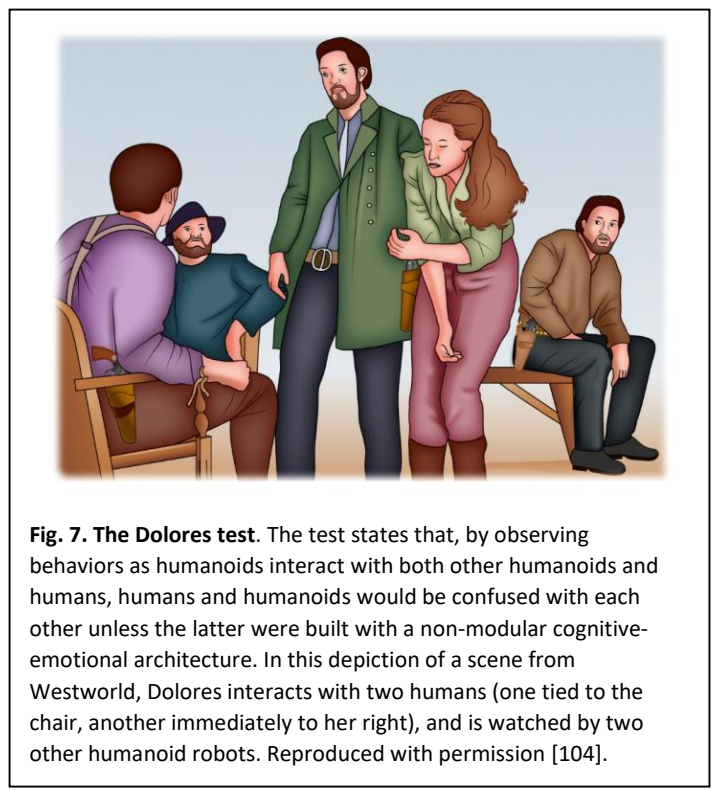

**Fig. 7. The Dolores test**. The test states that, by observing behaviors as humanoids interact with both other humanoids and humans, humans and humanoids would be confused with each other unless the latter were built with a non-modular cognitive-emotional architecture. In this depiction of a scene from Westworld, Dolores interacts with two humans (one tied to the chair, another immediately to her right), and is watched by two other humanoid robots. Reproduced with permission [104].

Of course, to make the test sufficiently precise, it would be necessary to operationalize "sophisticated behaviors." In particular, Turing believed that the question of whether or not machines can think was "too meaningless" to deserve discussion, and therefore sought to formulate the "Imitation Game," with the interrogator separated from the machine and the other person, as described above.

It has been argued that passing the Turing test does not necessarily show "intelligence." For example, the Chinese Room argument by Seale [100] points out that complex rule-based transformations may yield apparent understanding when only mechanical rules are being applied [101]. Could the behavior required to pass the Dolores test be generated by a set of complex rules? Although a more extended discussion should be taken up elsewhere in the future, the Chinese Room argument itself has fostered intense debate [101]. For instance, both the "systems reply" and the "virtual machine reply" argue that understanding is not present at the level of rule implementation per se (assuming it would be possible), but at the "systems level" including the rule-implementation mechanisms and (possibly all) other components at play – thus creating a "virtual" agent that does understand.

## 6. CONCLUSIONS

How should intelligent, autonomous robots be built? Information-processing architectures have mostly focused on cognitive components. In the past two decades, emotion components have been increasingly described and advocated as a way to improve human-robot communication, including the understanding of human facial expressions and the generation of expressions to communicate internal states. The survival argument has also been emphasized; emotion can then be used to help prioritize processes and actions that foster continued existence.

The central argument of the present framework is not that emotion is needed – the answer is a definitive "yes" – but that emotion and motivation need to be integrated with *all* information-processing components. This implies that cognitive-emotional integration is a principle of the architecture. Emotion is not an "add on" that endows a robot with "feelings" (for instance, reporting or expressing its internal state). It allows the *significance* of percepts, plans, and actions to be an integral part of all its computations.

Another implication of the present framework is that the traditional modular engineering approach is inadequate for autonomous robots. What we know about behavior and the brain of mammals (indeed of vertebrates more generally) suggests that an approach heavily influenced by biology is needed. In such framework, centralized control is abdicated, and distributed control is adopted. Cognitive and brain science have struggled with the concept of a homunculus, where "too much" intelligence is relegated to one of its components (but how does that component itself acquire its sophistication?) [102, 103]. Information-processing architectures for artificial agents need, likewise, to banish such notions of centralized control (a homunculus).

To return to the scenarios inspired by the Westworld series. Humans cannot "lose all emotion" and continue with their daily lives. I content that sophisticated, autonomous robots will prove the same. They will need cognition and emotion as integrated "components" of their information-processing architecture.

Acknowledgements
I am grateful to the National Institute of Mental Health for continued research support (R01 MH071589 and R01 MH112517) and Heiko Neumann for feedback on the manuscript. A reviewer of the manuscript also provided excellent feedback. I thank Christian Meyer and Anastasiia Khibovska for assistance with figures and references.also provided excellent feedback. I thank Christian Meyer and Anastasiia Khibovska for assistance with figures and references.

Figure captions

**Fig. 1. Information-processing architecture of intelligent robots**. (**A**) Modular architecture with separate cognitive (blue) and emotional (red) components. The cognitive module also inhibits the emotion module when deemed necessary. (**B**) Non-modular architecture with integrated components. Reproduced with permission [104].

**Fig. 2. Large-scale anatomical connectivity**. (A) Network analysis of anatomical datasets reveals massive interconnectivity among all sectors of the brain. (B) Illustration of the major lobes of cortex (top) and some of the regions mentioned in the text (bottom). Part (A) reproduced with permission [22].

**Fig. 3. Overlapping networks**. In standard clustering approaches, communities are comprised on disjoint sets of nodes. In the framework of overlapping networks (or mixed membership), each brain region affiliates with multiple communities with varying strengths called "membership values." For each brain region, these are probability-like values between 0 and 1 that sum to 1 (summed across communities). Here, the membership values of region $R_i$ are based on strengths to each of communities $C_1$ to $C_4$. Reproduced with permission [55].

**Fig. 4. Cognition and emotion in information-processing architectures**. (**A**) Central components of existing cognitive architectures, with a modular and largely hierarchical organization. (**B**) Emotion can be considered as an additional module of the architecture that interfaces with some of the cognitive components. In particular, it can be used to determine "mood" and "feeling". The region with dashed outline corresponds to the architecture in part A.

**Fig. 5. Dual competition model**. (**A**) Dual competition framework where emotion and motivation are integrated with perception and cognition. (**B**) Schematic representation of some of the network interactions linked to the model. In particular, the valuation network is important in the determination of affective and motivational significance. Reproduced with permission [105].

**Fig. 6. Cognitive-emotional integration**. (**A**) A priority map contains spatial locations of behavioral importance which are determined by combining multiple factors, including emotional/motivational ones. (**B**) Perceptual competition is influenced by multiple factors which determine the stimulus representation that should to be prioritized ("wins the competition"). (**C**) Executive control can be conceptualized as a series of processes that are partly interdependent (here, inhibition, updating, and shifting). Affective and motivational inputs engage the same processes at the core of cognition. (**D**). Behavioral performance is a function of the intensity of the emotional stimulus or context.

**Fig. 7. The Dolores test**. The test states that, by observing behaviors as humanoids interact with both other humanoids and humans, humans and humanoids would be confused with each other unless the latter were built with a non-modular cognitive-emotional architecture. In this depiction of a scene from Westworld, Dolores interacts with two humans (one tied to the chair, another immediately to her right), and is watched by two other humanoid robots. Reproduced with permission [104].